\documentclass[10pt,twocolumn]{article}

\usepackage{cvpr}
\usepackage{times}
\usepackage{graphicx}
\usepackage{amsmath}
\usepackage{amssymb}
\usepackage[caption=false]{subfig}
\usepackage[export]{adjustbox}
\usepackage[pagebackref=true,breaklinks=true,colorlinks,bookmarks=false]{hyperref}

\cvprfinalcopy 

\ifcvprfinal\fi
\begin{document}

\title{Dual Embedding Expansion for Vehicle Re-identification}

\author
{$^*$Clint Sebastian$^{1, 2}$, $^*$Raffaele Imbriaco$^{1}$, Egor Bondarev$^{1}$, Peter H.N. de With$^{1, 2}$\\
$^1$VCA Group, Eindhoven University of Technology  \hspace{2mm} $^2$Cyclomedia B.V \\
$^*$ \textit{\small equal contribution} \\
{\tt\small \{c.sebastian, r.imbriaco\}@tue.nl}}


\maketitle

\begin{abstract}
    Vehicle re-identification plays a crucial role in the management of transportation infrastructure and traffic flow. However, this is a challenging task due to the large view-point variations in appearance, environmental and instance-related factors. Modern systems deploy CNNs to produce unique representations from the images of each vehicle instance. Most work focuses on leveraging new losses and network architectures to improve the descriptiveness of these representations. In contrast, our work concentrates on re-ranking and embedding expansion techniques. We propose an efficient approach for combining the outputs of multiple models at various scales while exploiting tracklet and neighbor information, called dual embedding expansion (DEx). Additionally, a comparative study of several common image retrieval techniques is presented in the context of vehicle re-ID. Our system yields competitive performance in the 2020 NVIDIA AI City Challenge with promising results. We demonstrate that DEx when combined with other re-ranking techniques, can produce an even larger gain without any additional attribute labels or manual supervision. 
\end{abstract}

\section{Introduction}

Large-scale traffic video analysis can enable efficient management of transportation infrastructure and traffic flow. With the availability of a large number of sensors, an Intelligent Transportation System (ITS) can be developed to facilitate AI-powered smart cities.
It is beneficial to periodically recognize a vehicle across different locations and cameras to estimate the traffic flow. This concept is known as vehicle re-identification (re-ID), where the objective is to match a specific vehicle irrespective of location, time or camera view. In essence, vehicle re-ID is a constrained image retrieval task. Given a query image, we rank all the database images based on their similarity to the query. Image retrieval systems perform two fundamental processes: retrieval, and re-ranking. First, a feature extractor produces a compact representation of the image to facilitate retrieval. Then a similarity score is computed for each representation. Second, a re-ranking technique is applied to indicate the relevance of the retrieved results.

\begin{figure}
    \begin{center}
        \subfloat{
        \includegraphics[width=0.95\linewidth,valign=t]{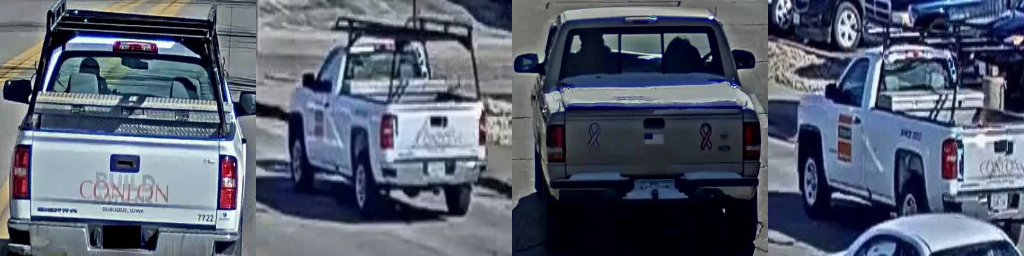}
        }\\
        \vspace{-11pt}
        \subfloat{
        \includegraphics[width=0.95\linewidth,valign=t]{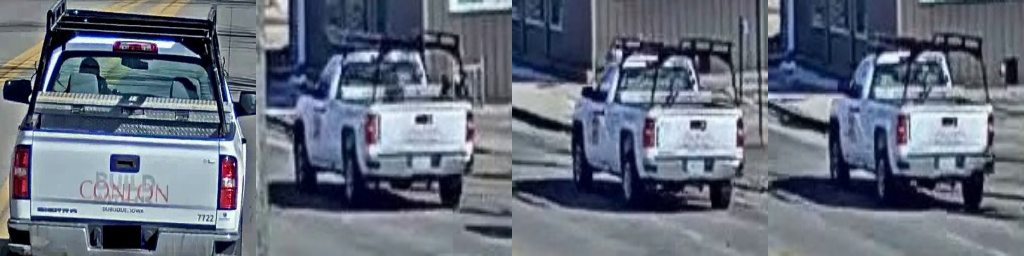}
        }
        \vspace{-8pt}
        \subfloat{
        \includegraphics[width=0.95\linewidth,valign=t]{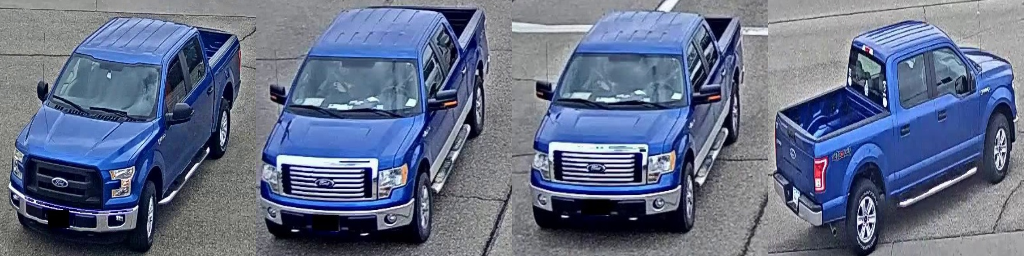}
        }
        \vspace{-11pt}
        \setcounter{subfigure}{1}
        \subfloat{
        \includegraphics[width=0.95\linewidth,valign=t]{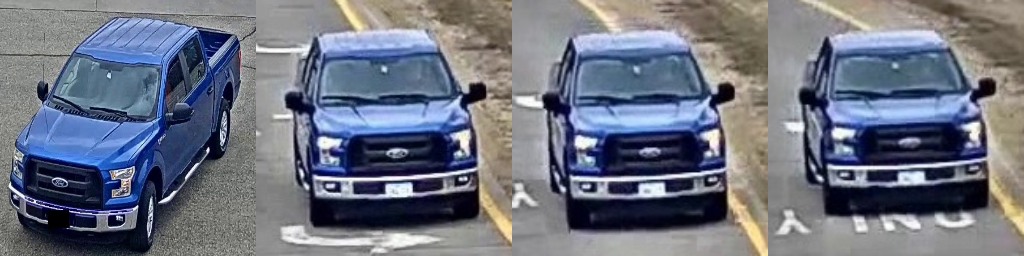}
        }
    \end{center}
       \caption{Qualitative comparison of top-$k$ ranks on the retrieved results (row 1, 3) and after applying the re-ranking schemes (row 2, 4).}
    \label{fig:short}
\end{figure}

Similar to other re-ID tasks such as person re-ID, vehicle re-ID also suffers from occlusion and low-quality images. However, vehicle re-ID is more challenging, since vehicles have low intra-class variations. For example, the same model may have only slight appearance changes (rims, lights) depending on their make. More drastic appearance variations such as glare can occur due to weather conditions, surroundings or camera position. Furthermore, a human body can be separated into a few semantic parts, while the outcome of separation of a vehicle into parts depends on the point of view. The front is symmetric with easily identifiable elements, such as the headlights, grille, and bumper, whereas the side is asymmetric and contains other parts like rims and doors. The combination of these factors makes the vehicle re-ID a more challenging task. For vehicle re-ID, license plates can be exploited to improve performance~\cite{tang2018single}. However, they are not always available at both sides or not visible, due to poor lighting conditions and occlusions. Apart from these issues, the usage could raise privacy concerns~\cite{uittenbogaard2019privacy}.
Besides license plates, other attributes can be generated to improve performance such as color, brand and vehicle type~\cite{Tan2019MulticameraVT,Chen2019MulticameraVT}. In this research, we focus on improving performance by producing a descriptive representation using nearest neighbors. Therefore, we propose a novel dual embedding expansion method offering promising results on the 2020 NVIDIA AI City Challenge. Our contributions are summarized as follows.
\begin{itemize}
    \item We propose an efficient embedding expansion strategy across CNN models and image scales that improves performance. The improved performance comes without any additional overhead during online retrieval. 
    \item We present an effective way to jointly use both tracklets and $k$-nearest neighbors ($k$-NN) of a query to enrich the embedding representation.
    \item We provide a comparative study of popular re-ranking techniques from landmark image retrieval on the CityFlow vehicle re-identification dataset. 
\end{itemize}

\section{Related Work}

\paragraph{Feature extraction.}
In recent years, vehicle re-ID has received large interest from the research community due to advances in deep learning. This has resulted in the publication of larger and more challenging datasets, such as CityFlow~\cite{Tang2019CityFlowAC}, VeRi-776~\cite{liu2016deep}, VehicleID~\cite{Liu2016DeepRD}, and VeRi-Wild~\cite{Lou2019VeRiWildAL}. Conventional feature extraction for image retrieval has been performed using hand-engineered descriptors e.g. SIFT~\cite{lowe2004distinctive}, SURF~\cite{Bay2006SURFSU} or HoG~\cite{dalal2005histograms}. However, modern approaches deploy Convolutional Neural Networks~(CNNs) as feature extractors. A major common focus has been on improving the descriptive capabilities of CNNs. Liu~\emph{et al.}~\cite{Liu2016DeepRD} propose to progressively refine the retrieval results using a combination of hand-engineered features (color, texture) together with global descriptors. Similarly, in~\cite{Tang2017MultimodalML}, features are aggregated with Bag-of-Words~\cite{sivic2003video} and are fused with CNN descriptors. 

Popular choices for CNN-based architectures for re-ID are generally part-based. Part-based models~\cite{liu2018ram,chen2019partition, AyalaAcevedo2019VehicleRP, Chen2019MulticameraVT} split the output of the network into several regions that learn part-specific features. These features are then merged into a single representation that is used as a final embedding for retrieval. Attentive models in~\cite{Tan2019MulticameraVT,Huang2019MultiViewVR,Khorramshahi2019AttentionDV,khorramshahi2019dual,teng2018scan,zhang2019part} train specialized modules to detect salient regions and improve retrieval performance. Attention modules are adapted for saliency in the spatial, channel, or temporal domain. Typically, CNNs are trained with a classification loss. However, metric learning methods such as triplet, contrastive, or center losses are almost ubiquitous in re-ID literature~\cite{liu2018ram, wen2016discriminative}. These are employed to improve performance by minimizing the distance between embeddings directly. Other angular losses, such as CosFace~\cite{Wang_2018_CVPR}, ArcFace~\cite{Deng_2019_CVPR}, and SphereFace~\cite{liu2017sphereface}, are also applied for re-ID tasks. 

Other approaches enrich the labels by adding additional information. Yan~\emph{et al.} in~\cite{yan2017exploiting} add brand, make and color to dataset instances and train a network with these extra attributes. This strategy is also adopted by in~\cite{Tan2019MulticameraVT}. Fine-grained labeling~\cite{He_2019_CVPR,zhao2019structural} with individual semantic parts such as `logo' and `windshield' are labeled and detected to improve the re-identification performance. In~\cite{wang2017orientation}, semantically relevant key points are annotated, instead of fine-grained attribute labels. A key disadvantage of these approaches is that this requires additional labeling. When additional data from a different domain is available, few approaches also apply domain adaptation strategies that are beneficial to improve performance\cite{liu2019supervised}.

\begin{figure*}
\begin{center}
\includegraphics[width=1.0\linewidth]{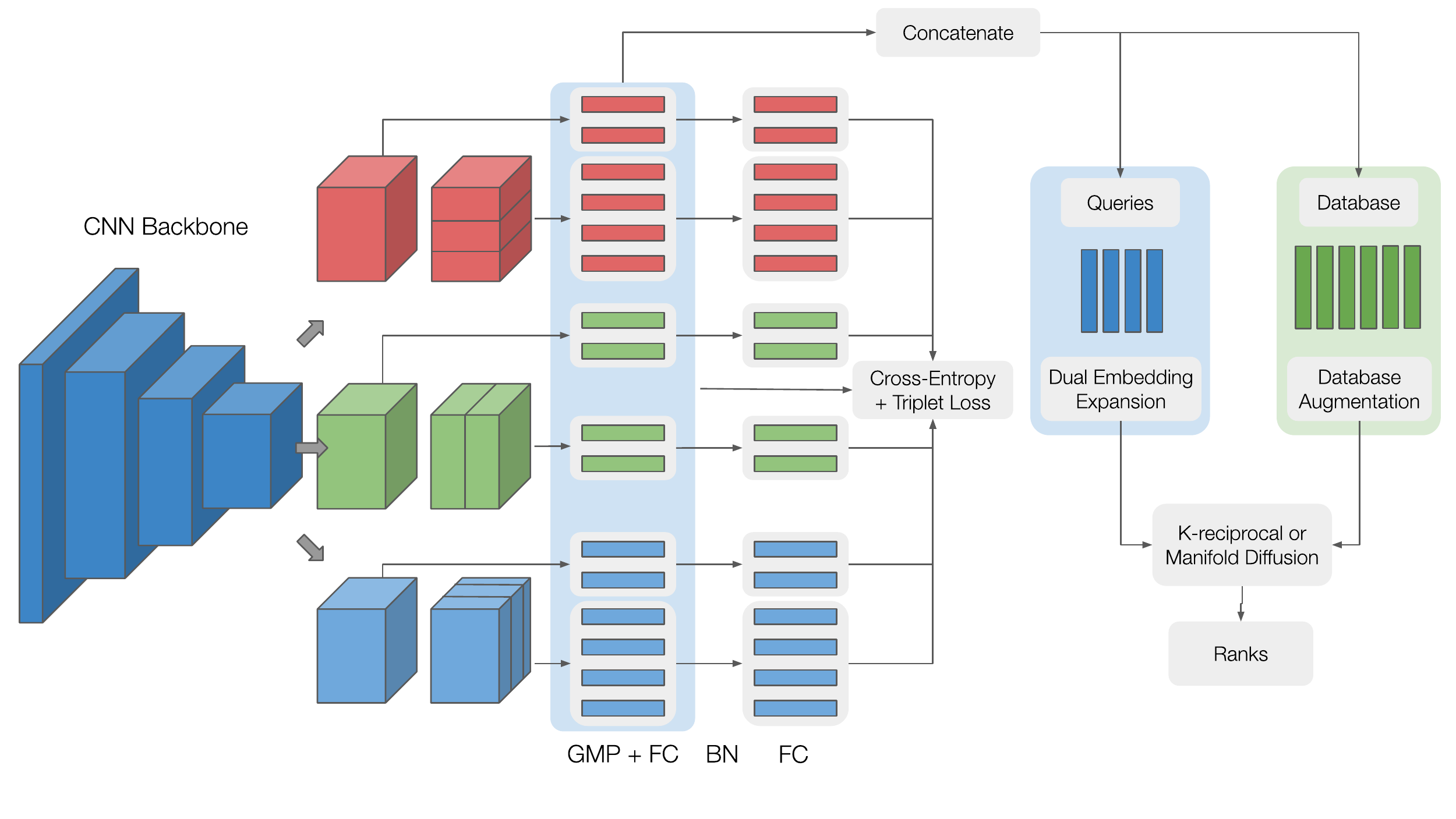}
\end{center}
   \caption{Overview of the feature extraction network and proposed re-ranking strategy. The features are extracted using a CNN backbone with a part-based model, similar to the Partition and Reunion Network~\cite{chen2019partition}. Each block represents partitions across height, channel, and width. All the outputs after global max-pooling (GMP) and fully connected layers are fed to triplet loss, whereas additional classification heads are applied with cross-entropy loss. Note that for simplicity, both losses are shown in a single block in the figure. The final output representations are extracted for both query and database/gallery images and then our dual embedding expansion is applied, followed by $k$-reciprocal re-ranking.}
\label{fig:system}
\end{figure*} 

\paragraph{Re-ranking.}
After obtaining the initial set of ranks from retrieval, a re-ranking step is usually utilized to refine the rankings. In vehicle re-ID~\cite{Huang2019MultiViewVR,Shankar2019ComparativeSO,Kanaci2019MultiTaskML,Chen2019MulticameraVT,chen2019partition,Zheng2019VehicleNetLR,liu2019supervised}, a common re-ranking technique is \textit{k}-reciprocal~\cite{Zhong2017RerankingPR}. It checks the \textit{k} nearest-neighbors of the query and their reciprocals in the database. The query representation is expanded by selecting top matches of the retrieved sets using the Jaccard distance.

A modification to $k$-reciprocal is presented in~\cite{huang2019deepfeature}. They employ the Mahalanobis distance when computing the similarity of the query and database images. Furthermore, they propose to re-organize the ranked list according to the vehicle tracklets. A similar technique is explored in~\cite{AyalaAcevedo2019VehicleRP}, with re-ranking by direct computation of the query similarity against the averaged tracklet representation. Huang~\emph{et al.}~\cite{Huang2019MultiViewVR} presents a novel re-ranking technique based on instance metadata. This technique trains a network to learn specific attributes like vehicle type, brand, color and uses the predictions to remove irrelevant matches. Afterwards, $k$-reciprocal is employed to improve the ranking list. Similarly, several other metadata-based constraints, such as speed, time, and location are exploited to improve vehicle re-ID performance \cite{Chen2019MulticameraVT,Tan2019MulticameraVT}.

Most of the previously discussed work uses the $k$-reciprocal encoding or metadata constraints for re-ranking. However, techniques common in image retrieval literature, such as query expansion and database augmentation, are largely absent. In~\cite{nguyen2019vehicle}, re-ranking is performed using spatial verification by handcrafted local descriptors and are aggregated using Bag-of-Words~\cite{sivic2003video}. Consistency across matches is enforced using visual words, and the local descriptors provide geometric constraints. Alternatively, in~\cite{Zheng2019VehicleNetLR}, the average query expansion is applied to enhance query representation, and database features are clustered in groups to the closest clustered centroid. In our work, we attempt to bridge the gap between vehicle re-ID and image retrieval, by studying the impact of popular re-ranking methods from image retrieval and proposing a novel dual embedding expansion technique that is suited for vehicle re-ID.

\section{Methods}
The proposed approach primarily consists of two parts, a CNN-based feature extractor for retrieval and post-processing using re-ranking techniques.

\subsection{Feature extractor}

Image embeddings are generated using a part-based model similar to the Partition and Reunion Network~\cite{chen2019partition} (PRN). Our model is composed of a convolutional backbone and three branches. Each branch encodes global information and partitions the height, width, and channel to obtain local information. Each branch does not share weights with the others, and partitions are mutually exclusive. However, there are minor differences with the PRN. We generate three horizontal or vertical partitions across the spatial dimensions and two partitions across the channel dimensions.  The output of each branch is pooled using global max pooling, followed by a fully connected layer to reduce dimensionality to 256. This is followed by a BNNeck~\cite{luo2019bag} and a classification head for each branch. We use a part-based network, due to its good performance for re-ID tasks. The number of partitions are reduced to avoid the design of excessively large descriptors. During test time, the outputs prior to the last convolutional layers are concatenated to obtain a single representation that encodes both global and local features. The resulting descriptor dimensions are 256~$\times$~17. For the final submission, we use the ResNet50~\cite{he2016deep},  ResNet50 IBN-a~\cite{pan2018two}, SE-ResNeXt50~~\cite{xie2017aggregated} and EfficientNet~\cite{tan2019efficientnet} architectures as the CNN backbones. The backbone selection and part-based model are motivated by their high performance in other retrieval, re-ID and classification tasks~\cite{huang2019deepfeature,chen2019partition,wang2018learning}. 

\paragraph{Loss functions}
For training, we use a combination of the cross-entropy loss with label smoothing regularization and the triplet loss with batch-hard negative mining. Cross entropy with label smoothing is given as 
\begin{equation}
    \mathcal{L}_{id} = \sum_i^N t_i \cdot \text{log}(y_i),
\end{equation}
where $y_i$ is the output for identity $i$ and label smoothing is applied to produce $t_i$  
\begin{equation}
    t_i = \begin{cases}
          1 - \frac{N-1}{N}\cdot\epsilon, \text{if} \hspace{2mm} i=j \\
          \frac{\epsilon}{N}, \text{otherwise},
          \end{cases}
\end{equation}
where $\epsilon$ is the regularization term (set to 0.1) and $N$, the number of classes.
The triplet loss is given as 
\begin{equation}
    \mathcal{L}_{triplet} = \sum_{a, p, n} \text{max}[D_{a,p} - D_{a,n} + m, 0],
\end{equation}
where $D_{a,p}$, $D_{a,n}$ are the distance between anchor ($a$) with positive ($p$) and negative ($n$), respectively. The triplet margin is $m$, the positive samples share the same class as the anchor and negatives are elements from different classes.
A key difference compared to previous approaches is that we apply triplet loss to each of the outputs of a fully connected layer, instead of a few selected outputs. We observe faster convergence and slightly improved performance with this approach. An overview of the system is shown in Figure~\ref{fig:system}.

\subsection{Re-ranking techniques}
 For re-ranking, a few techniques are considered. After inference, each embedding from different networks has the same size of 256~$\times$~17 dimensions. For similarity between the representations, cosine similarity is applied. We have selected the re-ranking techniques by their prevalence in image retrieval and re-ID literature. 
 
\paragraph{Dual embedding expansion.} The dual embedding expansion (DEx) comprises of two parts. The first part combines representations across models and image scales in an efficient manner. Typically, most ensemble strategies in re-ID or retrieval concatenate the representations, which results in very large embeddings. Although this improves performance, it is computationally expensive during the similarity computation. Similar to different models capturing distinct features of the same image, image scales also capture intricate details of the same image. Given an image \textit{I} and a scale $s \in \mathcal{S}$, we pass each image $\textit{I}$ at scale $s$ to the model $m\in \mathcal{M}$. Therefore, we propose to expand the embedding to produce features $f_{ms}$ by computing

\begin{equation}
    f_{ms} = \sum_{m_i \in \mathcal{M}} \sum_{s_j \in \mathcal{S}} m_i(s_j({I})).
\end{equation}
The final embedding $f_{ms}$ is the $L_{2}$ normalized after averaging features from different models at multiple scales. 
The second part leverages both tracklet information and the $k$-NN to improve the representations. Given a query and the gallery of images, we extract $f_{ms}$ for each, denoted as $Q_{ms}$ and $G_{ms}$, respectively. Since most re-ranking techniques are sensitive to the initial set of rankings, we utilize the tracklet information to construct a new aggregated gallery $T_{ms}$. The new tracklet gallery $T_{ms}$ is constructed by averaging all the embeddings in a given tracklet. We compute the cosine similarity between $Q_{ms}$ and $T_{ms}$ to obtain the top tracklets for each query. The original elements of the tracklets are placed back at the tracklet ranks to obtain a new set of ranks. Following this, we apply $\alpha$ query expansion by utilizing the new ranks generated from the tracklet information. Therefore, for a given track $t$, we compute the newly sorted gallery $g_t$ and the renewed query $\hat{q}_{i}$ by
\begin{figure}[t]
    \begin{center}
    \includegraphics[width=\linewidth]{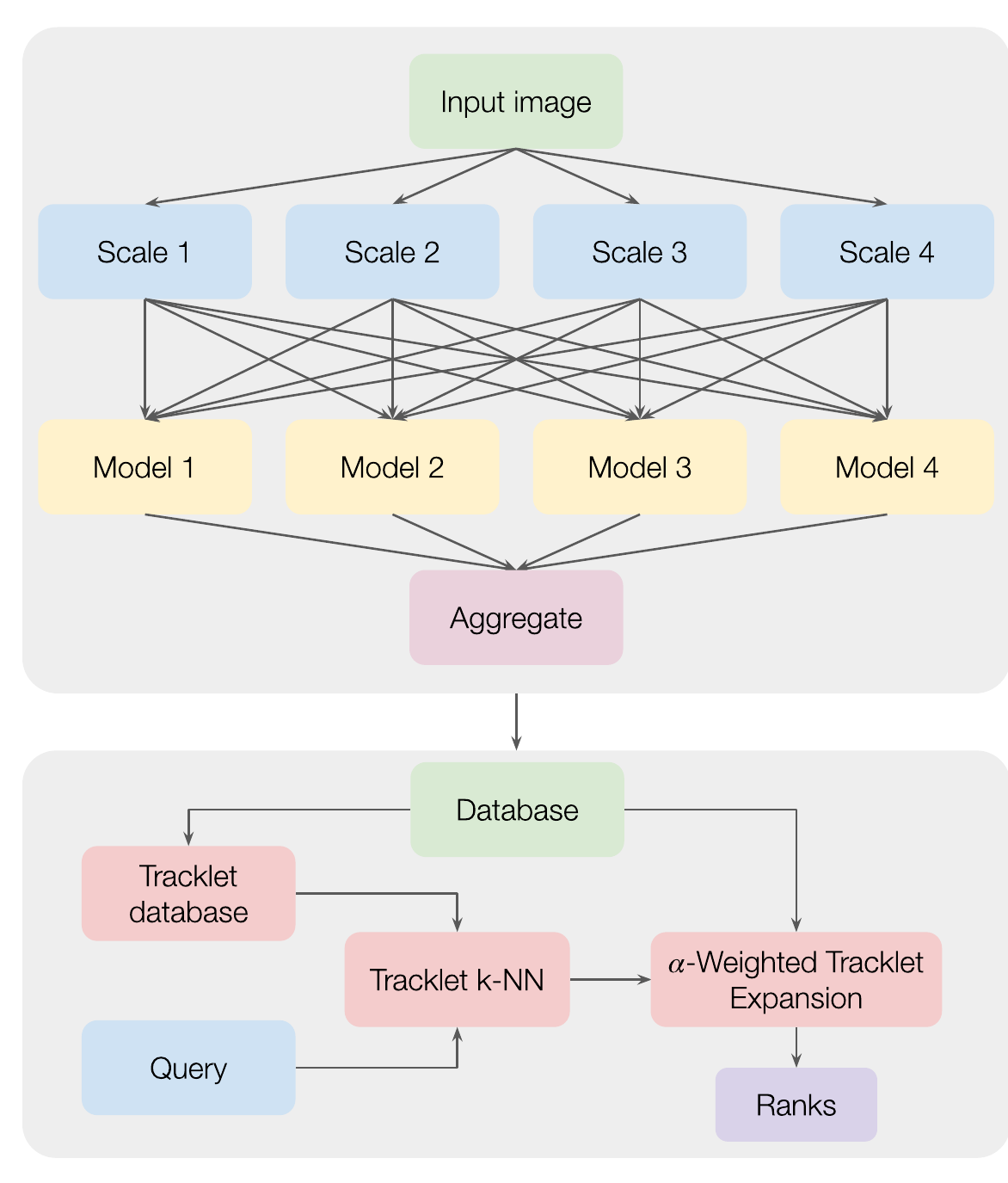}
    \end{center}
       \caption{Dual embedding expansion: The input image is processed at multiple scales and fed to all the models in a fully connected manner. The output representations are aggregated to produce an enriched embedding and are fed to tracklet-based query expansion to enhance the representation.}
    \label{fig:dex}
\end{figure}
\begin{equation}
    g_{t} = k\text{-NN}_{t}(q_{i}, g_{i}),
\end{equation}
\begin{equation}
    \hat{q}_{i} = q_{i} + \sum_{k} g_{t} \times {\text{cos}(q_i, g_t)^{\alpha}}, 
\end{equation}

\noindent where $q_i \in Q_{ms}$ and $g_i \in G_{ms}$.
The $k$-NN$_{t}$ and ``cos" refer to the $k$-nearest neighbors obtained through the tracklet information and cosine similarity, respectively. The $g_t$ is the sorted gallery based on the tracklets for a query $q_i$. The parameters $k$ and $\alpha$ denote the number of top-$k$ matches and power to weight the distance, respectively. A key advantage of the method is that the new representation retains the same size as the single-model, single-scale representation. Compared to other ensemble strategies such as concatenation, this reduces online computation costs while calculating similarity. The similarity computation cost is the same as a single-model, single-scale representation.
\vspace{-2mm}
\paragraph{Database-side feature augmentation.} We also perform database-side feature augmentation (DBA) by constructing a $k$-NN graph of the database to aggregate features. Instead of relying on tracklet information, only nearest neighbors are utilized. This allows to control the number of aggregated gallery images and ensures that all embeddings are expanded using the same amount of data. DBA commonly increases performance in landmark retrieval systems when there are occlusions or visibility constraints~\cite{Arandjelovic2012ThreeTE}.

\paragraph{Re-ranking with $k$-reciprocal encoding.}
In $k$-reciprocal encoding, a robust set of matches is constructed given an initial ranking. Given the $k$-NN neighbours of query $q_i$, the $k$-reciprocals of $q_i$ in $\mathcal{R}(q_i, k)$ are defined in~\cite{Zhong2017RerankingPR} as
\begin{equation}
    \mathcal{R}(q_i, k) = \{g_i | (g_i \in k\text{-NN}(q_i)) \land (q_i \in k\text{-NN}(g_i))\}. 
\end{equation}
In order to include positive images that may have been left out of the top-$k$ ranks, the new set $\mathcal{R}^*(q_i, k)$ is constructed by iteratively adding the half of the $k$-reciprocal images. These images are only added if the following condition holds
\begin{equation}
    |\mathcal{R}(q_i, k) \cap \mathcal{R}(g_i, \frac{1}{2}k)| \geq \frac{2}{3}|\mathcal{R}(g_i, \frac{1}{2}k)|,
\end{equation}
which avoids the addition of too many negative samples to the re-ranked set. Afterwards, the Jaccard distance between the query and gallery reciprocal sets is computed, based on the assumption that images with similar reciprocal sets are closely related. The final representation is a weighted linear combination of the original distance and the Jaccard distance, using a weighting parameter $\lambda$.

\paragraph{Diffusion.} In contrast to previous methods, diffusion considers the complete dataset manifold and propagates image similarities through the adjacency matrix~\cite{Donoser2013DiffusionPF, Iscen2017EfficientDO}. In diffusion, random walks through the similarity graph are computed, to spread the query similarity across the full adjacency graph. The results from the iterative computation of these random walks is a ranking matrix that accounts for the intrinsic structure of the dataset manifold. The affinity matrix $A$ is symmetrically normalized by
\begin{equation}
    S = D^{-1/2} A \, D^{-1/2}
\end{equation}
with $D = \text{diag}(A \, \mathbf{1}_n)$ where $\mathbf{1}_n$ is the unity vector of size $n$. We follow the iterative procedure in~\cite{Iscen2017EfficientDO}. For a given initial $\mathbf{f}^0$ vector, we use an iterative scheme for diffusion given by,
\begin{equation}
    \mathbf{f}^t = \alpha \, S \,\mathbf{f}^{t-1} + (1 - \alpha)\,\mathbf{y},
\end{equation}

\noindent where $\mathbf{f}^t \in \mathbb{R}^d$, $S$ is the transition matrix, $\mathbf{y}$ an $\ell^1$ vector. The parameter $\alpha$ regulates the spread of the manifold structure of the affinity matrix to the query points. Parameter $t$ is the iteration count of the solver. Besides $\alpha$ and $t$, diffusion employs two additional parameters. The parameters $k, k_q$, are the local constraints of the affinity and similarity matrices to remove noise. A monomial kernel is used as in~\cite{Iscen2017EfficientDO}

\section{Experiments}
\subsection{Dataset and metrics}
The CityFlow-ReID dataset consists of 56,277~images of~666 different vehicle classes. The dataset is divided into two parts, a training set of 36,935~images of 333~vehicle classes from 1,897 video tracks, and a test set of 18,920~images of 333~vehicle classes from 798~video tracks. The remaining 1,052~images of the test split are used as the query or probe set. For validation, we leverage the training set by holding back a 100 identities as a validation set, while training on the 233~classes. For reporting, the mean Average Precision (mAP) and the Top-1 and Top-5 accuracy of the cumulative match curve (CMC@1, CMC@5) are used.

\subsection{Implementation details}

We have trained eight different CNN backbones, ResNet50 (+CBAM~\cite{woo2018cbam}), ResNet50-IBN-a (+CBAM), SE-ResNeXt50 and EfficientNet-B1, B2, B3 with the Adam optimizer~\cite{kingma2014adam}. In our experiments, we have found that larger EfficientNet and ResNet-based models are lowering the performance. Hence, the experiments are restricted to the smaller variants. The learning rate is set to $2\times 10^{-4}$ and decays at an exponential rate with exponent 0.01 at each epoch. Triplet and cross-entropy losses are employed and the models are trained for 500 epochs. Cross-entropy and triplet losses are weighted by a factor of 2 and 1, respectively. Label smoothing~\cite{szegedy2016rethinking} is utilized for the classification loss with $\epsilon = 0.1$. Image triplets are obtained using a batch-hard mining strategy as in~\cite{hermans2017defense}. Per batch, 8~classes are sampled containing between 4 to 8 images each, depending on the CNN backbone. The images are resized to $288 \times 288$ pixels, and augmentations such as horizontal flipping, contrast, and random erasing~\cite{zhong2017random}, are applied. 

We deploy both dual embedding expansion and database augmentation and use the closest 20 and 10~neighbors, respectively. The value of $\alpha = 2$.  This is followed by $k$-reciprocal encoding and its parameters are set to $\lambda = 0.5$, $k_1 = 60$ and $k_2 = 30$. 

\subsection{Ablation studies}
The ablations studies are provided over the baseline network (ResNet50-IBN-a). First, the benefits of adding additional synthetic data during training are explored. A total of 193 synthetic vehicle IDs are randomly chosen and added to the training set.  The IDs are added such that a relatively balanced ratio between synthetic and real images is maintained. Second, the impact of different re-ranking schemes are studied which involves query expansion, database side augmentation, diffusion and the $k$-reciprocal encoding. 

\paragraph{Impact of additional synthetic data.}
The fourth edition of the AI City Challenge includes, in addition to the train, test and query image sets, a collection of synthetic vehicle images. We study how the addition of synthetic data impacts the performance of our baseline network ResNet50-IBN-a.  Three distinct scenarios are considered. First, training with real images only. Second, learning using a balanced number of synthetic and real images. Third, we train by merging both image sets. Note that to ensure a fair comparison, the test set in our split is composed only by real images. The results from these experiments are shown in Table~\ref{table:extra_Data}.

\begin{table}[t]
    \centering
    \begin{tabular}{lccc}
    \hline
    \textbf{Training set} & \textbf{mAP} & \textbf{CMC@1} & \textbf{CMC@5} \\ \hline
    Real only & 68.0 & \textbf{89.7} & \textbf{92.3} \\
    Balanced &\textbf{ 72.4} & 87.9 & 91.4 \\
    Full & 68.1 & 87.7 & 91.6 \\ 
    \hline \\
    \end{tabular}
    \caption{Performance of ResNet50-IBN-a with different real to synthetic data ratios. Balanced refers to a 1:1 ratio.}
    \label{table:extra_Data}
    \vspace{-2mm}
\end{table}

For the second scenario, there is a data imbalance issue. 
We try to balance the number of real to synthetic images in a 1:1 ratio. 
However, in our experiments, this ratio is only an approximate due to the variable number of images per tracklet. In total, we consider 426 unique IDs, out of which 233 are from the real dataset. For the final submission, we consider a total of 526 classes (including 333 real) for training the models. It can be observed that balancing the number of real and synthetic samples provides better performance than using only real data. The balanced inclusion of synthetic images increases the baseline mAP by roughly 4.3\%, whereas training with all the synthetic samples only yields a minor gain. However, the CMC metric reduces slightly with the inclusion of synthetic data. We conjecture that the addition of the synthetic images acts as a form of regularization and prevents the network from overfitting. Nevertheless, the addition of too many samples from the synthetic domain may be changing the underlying dataset statistics, thereby reducing the quality of the learned embeddings. Therefore, further experiments are using the `Balanced' training set.

\paragraph{Impact of dual embedding expansion.} The dual embedding expansion has two components, a model and scale component and a tracklet-based query expansion. For a single model, we provide the impact of different scales in Table~\ref{table:scale}. Complementary features from different scales improve the performance without additional online similarity computation costs. As expected, the scale that is used to train offers the best performance. Any other scale reduces the performance. However, the aggregation of all the scales results in the highest gains.
\begin{table}[ht]
    \centering
    \begin{tabular}{cccc}
    \hline
    \textbf{Scale} & \textbf{mAP} & \textbf{CMC@1} & \textbf{CMC@5} \\ \hline
    0.9 & 69.3 & 87.2 & 90.4 \\
    1 & 72.4 & 87.9 & 91.4 \\
    1.1 & 71.6 & 87.7 & 91.5 \\
    1.2 & 70.8 & 88.2 & 91.4 \\
    All & \textbf{73.8} & \textbf{89.0} & \textbf{92.0} \\ 
    \hline \\
    \end{tabular}
    \caption{Impact of the image scale on re-ID performance.}
    \label{table:scale}
\end{table}

When combined with $\alpha$-weighted tracklet query expansion, the performance improves further. We compare the proposed DEx with other popular query expansion methods in~\ref{table:dex}. For fair comparisons, single-scale DEx is also compared. DEx consistently outperforms other query expansion techniques. As shown in Table~\ref{table:dex}, higher values for $k$ yields higher performance. In the final submission, we have set $k$=20 for DEx to prevent over-tuning on the validation set.


\begin{table}[ht]
\centering
\begin{tabular}{ccccc}
\hline
\textit{\textbf{k}} & \textbf{AQE} & \textbf{$\alpha$QE} & \textbf{DEx-\textit{ss}} & \textbf{DEx-\textit{ms}} \\ \hline
5                   & 74.6        & 74.5         & 76.6           & 77.7           \\
10                  & 75.4        & 75.3         & 77.1           & 78.0           \\
20                  & 76.6        & 76.6         & 77.6           & 78.4           \\
40                  & 77.1        & 77.2         & 77.9           & 78.5           \\ \hline
\end{tabular}
\vspace{2mm}
\caption{Comparison of retrieval performance (mAP) with different query expansion techniques. DEx-$\textit{ss}$ and DEx-$\textit{ms}$ are dual embedding expansion with single and multi-scale features.}
\label{table:dex}
\end{table}
\vspace{-3mm}

\paragraph{Impact of database augmentation.}  Previous work~\cite{chen2019partition} has also explored a similar technique by encoding the entire database tracklets as the average representation of its images. Whereas this limits the possibility of polluting the database representations with those of negative IDs, it prevents the expansion with images belonging to other tracklets. In our work, we compute the $k$-NN of each database descriptor and aggregate the top-$k$ matches as the new image representation. 
\begin{table}[ht]
    \centering
    \begin{tabular}{cccc}
    \hline
    \textbf{\textit{k}} & \textbf{mAP} & \textbf{CMC@1} & \textbf{CMC@5} \\ \hline
    5 & 76.0 & 86.7 & 91.2 \\
    10 & 76.9 & 86.1 & 91.9 \\
    20 & 78.1 & \textbf{87.0} & \textbf{92.0} \\
    40 &\textbf{ 79.2} & 86.7 & 91.7 \\ \hline \\
    \end{tabular}
    \vspace{-2mm}
    \caption{Retrieval performance with DBA for various values of $k$.}
    \label{table:dba}
    \vspace{-5mm}
\end{table}
This provides additional flexibility when re-ranking, since the value of $k$ can be tuned for the best performance. These results are presented in Table~\ref{table:dba}. In general, increasing the number of neighbors aggregated into the database representation has a beneficial effect on the mAP metric. However, the CMC metric shows a slight deterioration. This indicates that some erroneous matches are being placed at the top-1 and top-5 ranks with DBA re-ranking, but correct matches are being ranked at adjacent positions. It can be observed that the highest overall mAP is obtained when $k=40$. Nevertheless, setting $k$ to this value lowers the performance on the private test set. This may occur if the test database has many visually similar instances of different vehicles. In this case, using too large values of $k$ reduces the discriminating power of the descriptors, so that $k=10$ is adopted in the final submission.

\paragraph{Impact of diffusion.} 
Table.~\ref{table:diffusion} contains the results from our experiments with diffusion. We observe that large values of $k$ or $k_q$ have a detrimental effect on retrieval performance. We set $\alpha=0.95$ and run for up to 25 iterations. Diffusion provides a gain of 8-10 mAP over baselines depending on the choice of $k$ and $k_q$.

\begin{table}[ht]
    \centering
    \begin{tabular}{ccccc}
    \hline
    \textbf{\textit{k}} & \multicolumn{1}{l}{\textbf{\textit{$k_q$}}} & \textbf{mAP} & \textbf{CMC@1} & \textbf{CMC@5} \\ \hline
    25 & 25 & 80.1 & 90.5 & 92.0 \\
    50 & 25 & 82.2 & \textbf{92.2} & 93.6 \\
    100 & 25 & 81.4 & 91.4 & \textbf{93.8} \\ \hline
    25 & 50 & 80.6 & 90.7 & 91.9 \\
    50 & 50 & \textbf{82.6} & 92.1 & 93.7 \\
    100 & 50 & 81.4 & 91.4 & 93.8 \\ \hline
    \end{tabular}
    \vspace{2mm}
    \caption{Retrieval performance with diffusion for various parameter settings.}
    \label{table:diffusion}
\end{table}

\begin{figure*}
    \begin{center}
        \subfloat{
        \includegraphics[width=0.95\linewidth,valign=t]{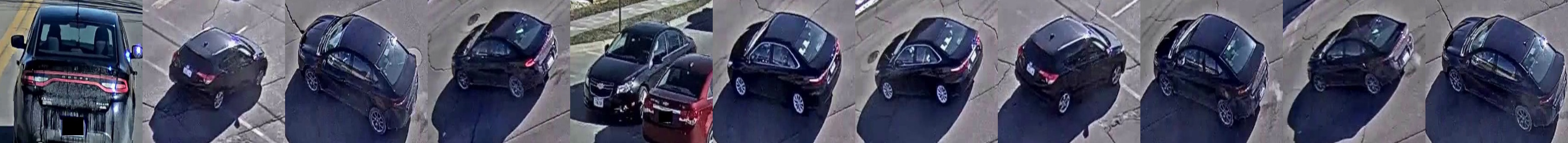}
        }\\
        \vspace{-11pt}
        \subfloat{
        \includegraphics[width=0.95\linewidth,valign=t]{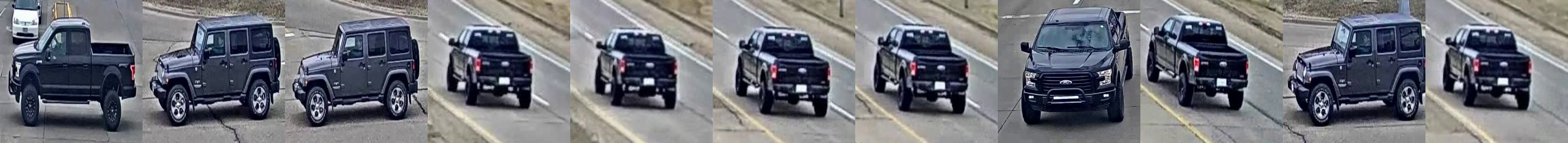}
        }
        \vspace{-11pt}
        \setcounter{subfigure}{0}
        \subfloat[][Top-10 retrieval results of the ResNet50-IBN-a network.]{
        \includegraphics[width=0.95\linewidth,valign=t]{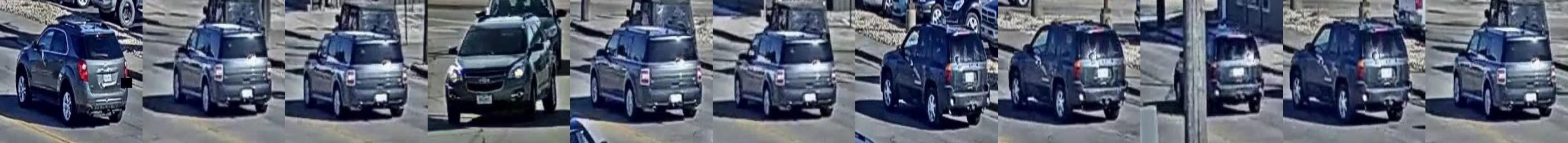}
        }

        \subfloat{
        \includegraphics[width=0.95\linewidth,valign=t]{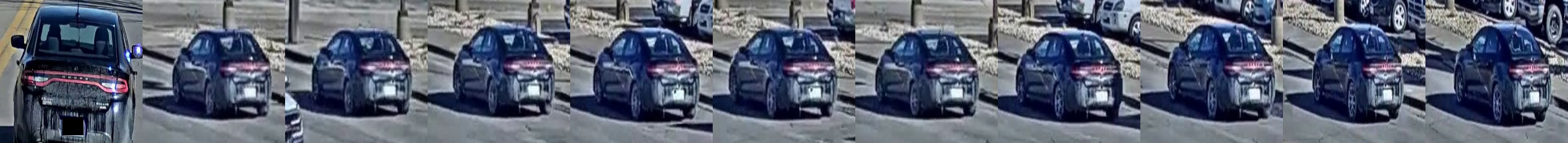}
        }\\
        \vspace{-11pt}
        \subfloat{
        \includegraphics[width=0.95\linewidth,valign=t]{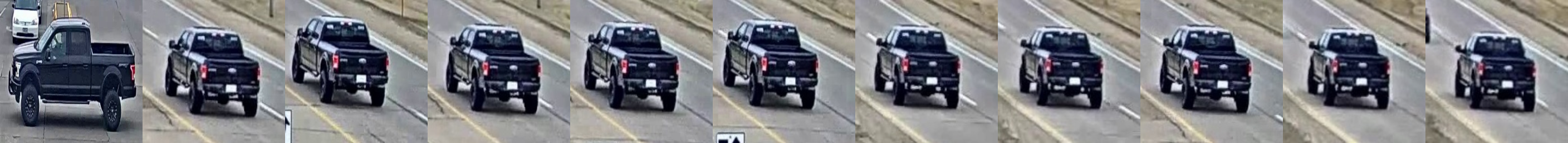}
        }
        \vspace{-11pt}
        \setcounter{subfigure}{1}
        \subfloat[][Top-10 retrieval results after re-ranking using DEx+DBA+$k$-reciprocal scheme.]{
        \includegraphics[width=0.95\linewidth,valign=t]{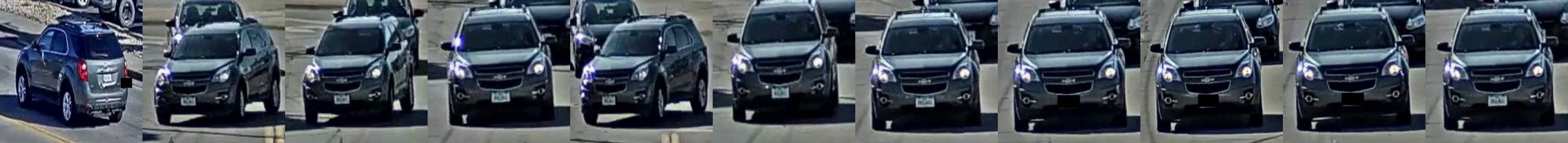}
        }

    \end{center}
       \caption{Retrieval results of the baseline ResNet50-IBN-a network (top) and re-ranking (bottom). Re-ranking removes possible mismatches (second row) and increases the homogeneity of the top ranks. The query is the leftmost image.}
    \label{fig:top10}
\end{figure*}

\paragraph{Impact of re-ranking schemes.} We also present a comparison of the studied re-ranking techniques in Table~\ref{table:combined}. Among these, the single best performing one is Diffusion, yielding an increase in mAP over the baseline of roughly 10 mAP points, outperforming other singleton methods. Nonetheless, DEx obtains CMC scores higher than others by roughly 1.1 and 0.8 points. The addition of tracklet information along with the combination of post-processing steps did not yield gains on the validation set. However, we observe up to 2\% improvement on the private test set. Figure~\ref{fig:top10} shows a few examples of retrieved vehicles in the test set using the baseline network and after re-ranking with our approach. The gain from diffusion similarly translates when combined with embedding enhancing techniques such as DEx and DBA.

\begin{table}[ht]
    \centering
    \begin{tabular}{lccc}
    \hline
    \multicolumn{1}{l}{\textbf{Re-ranking scheme}} & \textbf{mAP} & \textbf{CMC@1} & \textbf{CMC@5} \\ \hline
    \multicolumn{1}{l}{Baseline} & 72.4 & 87.9 & 91.4 \\
    +DEx & 78.4 & \textbf{93.3} & \textbf{94.7} \\
    +DBA & 76.9 & 86.1 & 91.9 \\
    +$k$-reciprocal & 80.4 & 91.8 & 93.9 \\ 
    + Diffusion & 82.2 & 92.2 & 93.6\\
    +Tracklet & 76.7 & 87.9 & 88.0 \\
    \hline
    \multicolumn{1}{l}{+DEx+DBA} & 81.8 & 89.6 & 94.1 \\
    \multicolumn{1}{l}{+DEx+$k$-reciprocal} & 82.5 & 94.4 & 95.2 \\
    \multicolumn{1}{l}{+DEx+Diffusion} & \textbf{84.3} & \textbf{94.4} & \textbf{95.4} \\
    \hline 
    \multicolumn{1}{l}{+DEx+DBA+$k$-reciprocal} & 83.7 & 88.6 & 94.0 \\
    \multicolumn{1}{l}{+DEx+DBA+Diffusion} & \textbf{85.0} & \textbf{92.6} & \textbf{94.7} \\
    
    \hline \\
    \end{tabular}
    \caption{Retrieval performance of different re-ranking techniques on the validation set. }
    \label{table:combined}
\end{table}

\subsection{Results on 2020 AI City Challenge} 
The results when using the proposed methods are submitted to the re-ID track of 2020 AI City Challenge. We have obtained an mAP at top-100 ranks (mAP@100) of 51.66 without any additional labeling or supervision. This is a large improvement over the baseline mAP@100 of 26.3~\cite{Tang2019CityFlowAC}. The results are obtained with the DEx+DBA+$k$-reciprocal combination, followed by pulling in the tracklets. The  DEx+DBA+Diffusion yields slightly lower performance with 50.78 mAP. This is possibly due to overfitting of the diffusion parameters on the validation set.

\section{Conclusions}

In this paper, we propose a dual embedding expansion strategy that leverages multiple networks, scales along with weighted tracklet information to improve the re-ID performance. Our embedding expansion yields competitive performance to the de-facto standard in re-ID, outperforming conventional query expansion methods. We have performed ablations to study the impact of synthetic data during training and discovered that a balanced addition of synthetic data is beneficial. Furthermore, a detailed study is presented of several re-ranking techniques in image retrieval for vehicle re-ID. Although $k$-reciprocal is the common choice for re-ranking, diffusion offers competitive results with similar computational costs. Our evaluation of the different re-ranking techniques shows that $k$-reciprocal or diffusion is the best for re-ID, which can be further enhanced by combining it with other embedding expansion techniques without any additional annotations or manual supervision. 

\paragraph{Acknowledgements.} We thank NVIDIA Corp. for their grant of a Titan Xp GPU for research.

{\small
\bibliographystyle{ieee_fullname}
\bibliography{bibliography}
}

\end{document}